\documentclass{article}
\usepackage{spconf,amsmath,graphicx}

\graphicspath{{pdf/}{jpeg/}{figures/}}
\DeclareGraphicsExtensions{.pdf,.jpeg,.png}

\usepackage{tabularx}
\usepackage{makecell}
\usepackage{multirow}
\usepackage{flushend}
\usepackage{xcolor}

\usepackage{pifont}
\newcommand{\cmark}{\ding{51}}
\newcommand{\xmark}{\ding{55}}

\usepackage{amssymb}

\newcolumntype{Y}{>{\centering\arraybackslash}X}

\title{Human Pose Estimation from Ambiguous Pressure Recordings with Spatio-temporal Masked Transformers}

\name{Vandad Davoodnia, Ali Etemad}
\address{Department of Electrical and Computer Engineering \& Ingenuity Labs\\Queen's University\\Kingston, ON, Canada}

\begin{document}
\topmargin=0mm
\ninept

\maketitle

\begin{abstract}
Despite the impressive performance of vision-based pose estimators, they generally fail to perform well under adverse vision conditions and often don't satisfy the privacy demands of customers. As a result, researchers have begun to study tactile sensing systems as an alternative. However, these systems suffer from noisy and ambiguous recordings. To tackle this problem, we propose a novel solution for pose estimation from ambiguous pressure data. Our method comprises a spatio-temporal vision transformer with an encoder-decoder architecture. Detailed experiments on two popular public datasets reveal that our model outperforms existing solutions in the area. Moreover, we observe that increasing the number of temporal crops in the early stages of the network positively impacts the performance while pre-training the network in a self-supervised setting using a masked auto-encoder approach also further improves the results.  
\end{abstract}


\section{Introduction}
\label{sec:intro}
Human pose estimation is an important task in computer vision research with applications in healthcare \cite{liu2020simultaneously}, robotics \cite{zhou2019soft}, autonomous driving \cite{gu2019efficient}, and action recognition \cite{duan2022revisiting}. With recent advances in deep learning, researchers have developed optical and depth models to estimate human 3D models \cite{clever2020bodies} or their body joint locations with sub-pixel accuracy \cite{iskakov2019learnable}. However, despite the impressive performance of vision-based models, privacy demands and challenging aspects like severe occlusions during daily activities have given rise to \textit{non-vision} pose estimation techniques. Since most human activities rely on their contact with the surrounding environment, smart textiles have been proposed as a solution in numerous applications such as health care \cite{liu2020simultaneously} and identification \cite{davoodnia2019identity}. 

Currently, pressure-recording carpets and bed mattresses are available commercially. These systems have been designed to capture the pressure profiles and postures of subjects. Researchers have been able to use such systems to identify sleeping posture \cite{davoodnia2022estimating} and even 3D joint coordinates \cite{liu2020simultaneously} reliably using deep learning models. However, unlike vision-based approaches, these models often inherit complex architectures to account for the \textit{noisy} and \textit{ambiguous} nature of the data. For example, systems using smart carpets for pose estimation need to utilize the limited information from feet pressure distribution to estimate the joint locations in 3D space \cite{luo2021intelligent}. Similarly, in-bed pressure systems record artifacts caused by blankets, movement, or stretching of pressure sensors \cite{liu2022pressure}. Furthermore, pressure-based deep learning models can only be trained on datasets with limited diversity, or alternatively by synthetic data \cite{clever2020bodies}, unlike vision-based models that frequently rely on large-scale and in-the-wild datasets. As a result, detecting the full posture in 3D from pressure data is deemed a challenging problem with models having up to ten times more error compared to the vision-based solutions \cite{luo2021intelligent}.

In this paper, we propose a self-supervised strategy for learning the human pose from ambiguous pressure data using a temporal adaptation of ViTPose \cite{xu2022vitpose} (a vision transformer model originally developed for pose estimation from `single-frame' `images'). Our solution is able to achieve state of the art in two large-scale in-bed and smart carpet datasets. More specifically, we first pre-train the encoder of our network in a masked image reconstruction task with a warm-up strategy to leverage the benefits of the pre-trained ViT models. Then we train the network on pressure data using a combination of objective functions for pose estimation. We show the effectiveness of the self-supervised pre-training and network design by comparing our model's performance to previous works and state-of-the-art vision-based models on both datasets.

Our contributions can be summarized as follow:
(\textbf{1}) We propose a temporal variation of the ViT \cite{dosovitskiy2020image} for pose estimation, setting a new state of the art in temporal pose prediction from ambiguous pressure data.
(\textbf{2}) We show that pre-training the ViT in a masked auto-encoder framework can positively impact the performance on both datasets used in this study.

\section{Related Works}
\label{sec:related}
\textbf{Human Pose Estimation.}
Due to recent advances in computer vision and the collection of large-scale datasets, great improvements have been achieved in 2D human pose estimation \cite{sun2019deep,xu2022vitpose,xiao2018simple}. 
However, these models generally fail to predict 3D poses directly due to the underlying ambiguity of the 3D pose in one frame. As a result, most recent models rely on temporal information \cite{reddy2021tessetrack} or multi-camera setups \cite{iskakov2019learnable}. More recently, self-supervised learning strategies have emerged as a viable solution to reduce the pose-ambiguity in monocular images \cite{kundu2022uncertainty}. In a recent work, a 2D to 3D pose uplifting solution was developed, utilizing a self-supervised training strategy by reconstructing masked joints or frames in a sequence of poses \cite{shan2022p}. 
In another direction, bigger models was suggested to learn temporal information directly alongside spatial information, at the cost of computational resources \cite{reddy2021tessetrack}. 
Despite the impressive performance of these works, most approaches still rely on an accurate initial prediction of the 2D joints or textural information from RGB images, which is not available when using pressure recordings from smart textiles. To address the challenges of ambiguity in input pressure data, we exploit temporal information in our network by implementing a temporal variation of ViTPose \cite{xu2022vitpose}. Moreover, we address the lack of diversity and large-scale data by pre-training our network via a self-supervised training strategy.

\noindent \textbf{Pressure-based pose estimation.}
The development of cost-efficient and ubiquitous tactile sensing systems has allowed researchers to model human interactions and gestures using pressure recordings.
For instance, the patterns of hand grasp through pressure-sensing textiles have been explored in \cite{zhou2019soft}, while other works have developed multi-modal solutions for hand gesture detection using a combination of cameras and e-textiles \cite{li2019connecting}. 
Similarly, several works have explored virtual-reality and healthcare applications via human gait recognition \cite{li2021wearable} and posture detection \cite{davoodnia2019identity} using an array of piezoresistive pressure sensors embedded in a carpet. 
A recent study \cite{luo2021intelligent} addressed the challenging task of 3D human pose estimation using only feet pressure distributions using a deep network based on 3D convolutions. Given the limited information available in feet pressure and the ambiguity of its mapping into a 3D pose, they achieved an error of 20$cm$, which is almost ten times more than vision-based approaches \cite{xu2022vitpose}. 
In another work, to disambiguate in-bed pressure maps, generation of human-like figures from a pressure recording matrix via a pre-processing block before pose estimation was proposed \cite{davoodnia2022estimating}. In the same line of work, in \cite{clever2020bodies}, a synthetic in-bed pressure dataset was presented to address the limited diversity of available datasets, allowing network pre-training for better performance in real-life applications.

\section{Method}
\begin{figure*}[t]
\includegraphics[width=1\textwidth]{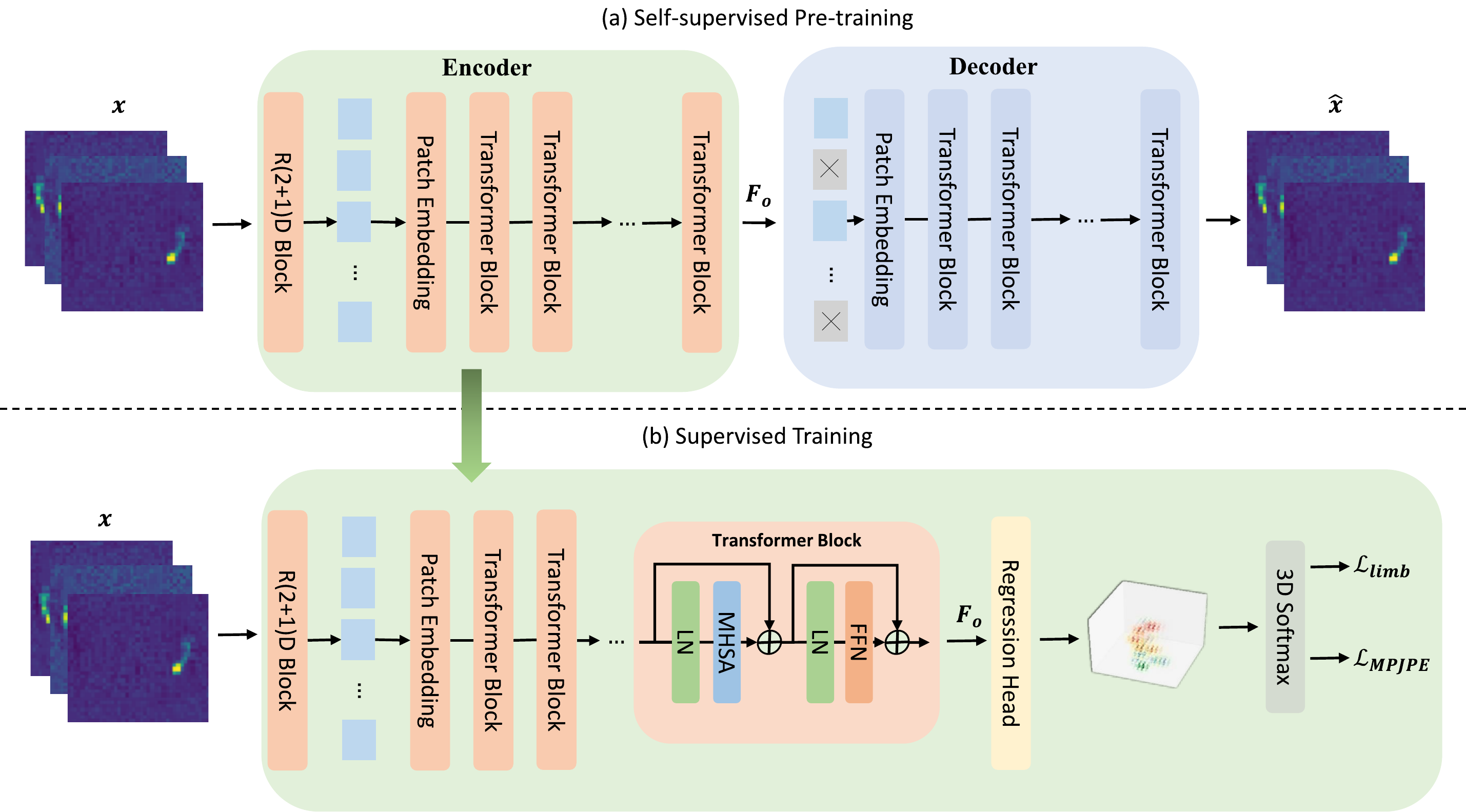}
\caption{An overview of our model is illustrated. (a) shows the masked auto-encoder, and (b) shows our pose estimation network.}
\label{fig:fig1}
\end{figure*}

\noindent \textbf{Problem setup.}
Let $X=\{ x_1,x_2,...,x_T \}, x_i \in \mathbb{R}^{WH}$ be a sequence of $T$ pressure distribution frames and $Y=\{ y_1,y_2,...y_T \}, y_i \in \mathbb{R}^{3J}$ be a sequence of $3$D ground truth joint locations. Our goal is to train a pose estimator encoder as illustrated in Figure \ref{fig:fig1}. We do so via learning the mapping of $\hat{Y}=P(Y|X,\theta,\theta_r)$, where $\theta$ and $\theta_p$ are the trainable parameters of the encoder and the regression head, respectively. We pre-train the encoder via self-supervision prior to pose estimation. The details of our network architecture, training steps, and implementation details are described in the following sections.

\noindent \textbf{Network Architecture.}
Similar to video-based vision transformers, we first tokenize the input frames $X$ into space-time cubes after applying a $(2+1)D$ convolutions \cite{tran2018closer}. Next, we embed the space-time cube tokens via a patch embedding layer, $F \in \mathbb{R}^{\frac{T}{d_t}\frac{W}{d}\frac{H}{d}C}$, and after a linear projection layer, pass them to an \textit{Encoder}, which is a ViT \cite{dosovitskiy2020image} pre-trained on ImageNet \cite{deng2009imagenet} followed by MS-COCO \cite{lin2014microsoft} datasets. Here, $d_c$ and $d$ are the number of temporal and spatial crops, respectively. The ViT is a sequence of transformer blocks, each consisting of multi-head self-attention (MHSA) and feed-forward network (FFN) layers. In our setting, the encoder operates on all of the input patches without using the masked tokens with an output of $F_o \in \mathbb{R}^{\frac{T}{d_t}\frac{W}{d}\frac{H}{d}C}$. As illustrated in Figure \ref{fig:fig1} (b), we adopt a simple \textit{Regression} head composed of two deconvolution layers and one $1\times1$ convolution layer on the reshaped $F_o$ following the common setting of the previous works \cite{he2022masked}. Specifically, each deconvolution block up-samples the reshaped feature vectors by a factor of two, and the $1\times1$ convolution layer predicts the joint location heatmaps for each keypoint in each frame $K \in \mathbb{R}^{T\frac{W}{4}\frac{H}{4}J}$.
The last module in our network is a \textit{decoder} $\hat{X}=D(X|F_o,Mask,\theta_d)$ used only during the self-supervised pre-training of our encoder to reconstruct the input patches from the encoder's output latent representations. We adopt a shallow transformer network to process $F_o$ and connect it to a linear projection layer to match the shape of the input patches. 

\noindent \textbf{Self-supervised pre-training.} 
Masked Auto-Encoders (MAE) for pre-training transformer networks has shown strong results on a variety of applications such as NLP \cite{devlin2018bert}, pose estimation \cite{xu2022vitpose}, and image classification \cite{he2022masked}. Inspired by this, we implement an MAE by masking the outputs of our encoder, $F_o$, and then passing the masked encoded features along with the masked tokens to our decoder. For this purpose, we adopt an asymmetric design for masked image reconstruction, where the encoder operates on fully observed data (without masked tokens), and the decoder is applied on the masked encoder output. Our encoder-decoder network reconstructs the masked patches by learning to predict the raw pixel values of each patch. We use MSE loss on the masked patches, excluding the unmasked patches, to train the network efficiently \cite{devlin2018bert}.

\noindent \textbf{Fine-tuning for Pose Estimation.}
In the fine-tuning step, we first freeze the decoder and warm up the regression head and patching block of the encoder for $2$ epochs. Then, we train all of the networks via minimizing the MSE loss between the predicted and ground-truth heatmaps, as well as the limb-length loss between the ground-truth and predicted keypoint coordinates, taken via SoftMax layer on each heatmap, by:
\begin{equation}
    L_{link}^i=
    \begin{cases}
      \hat{L}_i-L_{95\%}, & \text{if $\hat{L}_i > L_{95\%}$} \\
      L_{5\%}-\hat{L}_i,  & \text{if $\hat{L}_i < L_{5\%}$}  \\
      0 & \text{otherwise,}
    \end{cases} 
\end{equation}
where $L_{5\%}$ and $L_{95\%}$ represent the $5^{th}$ and $95^{th}$ percentile of each of the limb lengths in the given dataset.

\noindent \textbf{Implementation details.}
We use the same architecture proposed in ViTPose \cite{xu2022vitpose} for our encoder and only modify the patching layer by replacing the normal convolution operator with a $(2+1)D$ convolution block. In all cases, we start the training from pre-trained weights provided for ViTPose-B \cite{xu2022vitpose} in eligible layers. More specifically, we only use random initialization for the fully connected layers that have different sizes than the vanilla ViTPose due to the increased number of patches. Our decoder consists of 4 blocks of transformer networks with hidden dimensions of $256$. We initialize the Temporal ViTPose weights from the pre-trained model of MAE \cite{he2022masked}, and perform masked region reconstruction with a masking rate of $75\%$. We train the encoder-decoder pipeline with a learning rate of $1e^{-3}$ using AdamW optimizer \cite{reddi2019convergence} for $200$ epochs with a weight decay rate of $0.1$. In the next stage, we warm up the network by training the regression head, newly initilized layers, and the patching layer of ViT in a supervised task with a learning rate of $1e^{-3}$ for $2$ epochs. Finally, we use a learning rate of $2e^{-4}$ for training the network on the supervised pose estimation task for a total of $150$ epochs.

\section{Experiments and Results}
\label{sec:exp}

\subsection{Datasets}
\label{sec:dataset}

\noindent \textbf{Intelligent Carpet.} This dataset contains $15$ actions performed by $10$ subjects in two hour-long videos \cite{luo2021intelligent}. The $3$D pose is obtained by triangulating AI-generated $2$D joints from two webcams. Moreover, synchronized recordings of pressure maps via sensor-embedded carpets are provided for pose estimation, resulting in over 1,800,000 frames. Following the prior work, we train our model on $7$ subjects and evaluate on the remaining $3$.

\noindent \textbf{SLP.} The simultaneously-collected multi-modal lying pose (SLP) dataset \cite{liu2020simultaneously} is a collection of multi-modal data, namely RGB, LWIR, depth, and pressure maps, recorded from 102 subjects in home and 7 subjects in hospital settings. We only use the no-cover condition, and leave the thin and thick cover conditions out. As the dataset does not provide direct 3D annotation, we use the available 2D annotations and the corresponding depth of the joints of subsequent postures to train our model. Following the standard evaluation scheme of the in-home set, we train our models on the first 90 subjects and use the remaining 12 for testing.

\subsection{Performance Metrics}
\label{sec:metrics}
\textbf{MPJPE:} For the Intelligent Carpet dataset, we report mean-per-joint-position-error (MPJPE) as the performance metric in line with previous research \cite{luo2021intelligent}.

\noindent\textbf{PCKh:} To compare our method with previous works on the SLP dataset \cite{liu2020simultaneously}, we report the percentage of correct keypoints at $50\%$ of head limb length threshold (PCKh@0.5 \cite{lin2014microsoft}) using only 2D predictions.

\subsection{Benchmarks}
\label{sec:bench}
In order to evaluate our proposed model on the Intelligent Carpet dataset, we adapt and modify commonly used pose estimators, namely ResNet \cite{xiao2018simple}, UNet \cite{ronneberger2015u}, HRNet \cite{sun2019deep}, and ViTPose \cite{xu2022vitpose}. Specifically, we concatenate the frames in the channel dimension and change the first convolution layer accordingly. 
This adaptation follows the same strategy of previous work on the Intelligent Carpet dataset \cite{luo2021intelligent}. On the SLP dataset, we modify prior works to adjust them to temporal data and compare them with our approach. To train the benchmarks, we initialize the models using their pre-trained weights. Finally, for a fair comparison, we fine-tune them with empirically-tuned learning rates and the same number of epochs as our proposed model.

\begin{table}
\centering
\begin{center}
\caption{MPJPE (cm) on Intelligent Carpet ($\downarrow$ is better).}
\label{table:tb1}
\setlength
\tabcolsep{3pt}
\footnotesize
\centering
\resizebox{1\columnwidth}{!}{
\begin{tabularx}{\columnwidth}{c|YYYYYY}
\Xhline{2\arrayrulewidth}
\multirow{2}{*}{\textbf{Method}}      & \multicolumn{6}{c}{\textbf{Frames}}        \\ 
\cline{2-7}
                                      &    1       &   4       &      8       &     12    &   20      &   32      \\
\Xhline{2\arrayrulewidth}
ResNet18                              &   41.4     &   40.9    &      37.5    &   33.4    &   28.8    &   29.8    \\  
ResNet50                              &   32.5     &   31.1    &      31.2    &   30.7    &   29.8    &   29.4    \\ 
ResNet101                             &   38.3     &   42.1    &      37.8    &   32.6    &   31.7    &   30.1    \\ 
UNet                                  &   34.9     &   33.8    &      32.4    &   32.1    &   30.4    &   28.7    \\ 
HRNet-W32                             &   \underline{25.4}     &   \underline{25.6}    &      25.4    &   25.2    &   24.8    &   24.6    \\ 
3DCNN \cite{luo2021intelligent}       &   33.5     &   28.7    &      \underline{24.6}    &   23.1    &   19.8    &   22.4    \\   
ViTPose                               &   \textbf{24.3}     &   \textbf{24.6}    &      \textbf{23.7}    &   \underline{22.3}    &   21.7    &   21.2    \\        
T-ViTPose (\textbf{Ours}, $C$=~4, $H$)     &   N/A      &   28.9    &      28.4    &   22.4    &   \underline{19.4}    &   \underline{16.9}    \\ 
T-ViTPose+MAE (\textbf{Ours}, $C$=~4, $H$) &   N/A      &   28.4    &      27.5    &   \textbf{21.6}    &   \textbf{17.8}    &   \textbf{16.5}    \\ 
\Xhline{2\arrayrulewidth}
\end{tabularx}
}
\end{center}
\end{table}

\subsection{Results}
\label{sec:results}
In Table \ref{table:tb1}, we compare our method against the benchmarks when different numbers of temporal frames are given in the Intelligent Carpet dataset. We observe that changing the architecture of pose estimators as proposed in \cite{luo2021intelligent} does not benefit the performance of ResNet50, HRNet, and UNet as much as 3DCNN and our proposed model. Furthermore, we observe that although ResNet101 has more parameters than the other architectures, it performs worse, hinting at over-fitting and the limited diversity of foot pressure distributions caused by data ambiguity. Furthermore, we show that only changing the cropping strategy and the first convolution layers of the ViTPose, significantly improves the performance of the pose estimator and achieves the lowest error among others. Finally, we show that by utilizing our self-supervised pre-training strategy, we can consistently improve the performance, where a $0.4mm$ reduction in error is observed at $32$ frames of pressure data. We also observe a similar effect in Table \ref{table:tb2}, where we compare the performance of our proposed solution to our implementations of previous research. We see that our approach is able to consistently achieve the best results where $12$ or more frames are available. We also see that self-supervised pre-training of our Temporal ViTPose consistently improves the performance.

\begin{table}
\centering
\begin{center}
\caption{{\small{PCKh@0.5}} on SLP (\textit{uncovered}) ($\uparrow$ is better).}
\label{table:tb2}
\setlength
\tabcolsep{3pt}
\footnotesize
\centering
\resizebox{1\columnwidth}{!}{
\begin{tabularx}{\columnwidth}{c|YYYYYYY}
\Xhline{2\arrayrulewidth}
\multirow{2}{*}{\textbf{Method}}    & \multicolumn{6}{c}{\textbf{Frames}}        \\ 
\cline{2-7}
                                      &    1       &       4   &      8       &     12    &   20      &   32      \\
\Xhline{2\arrayrulewidth}
ResNet50 \cite{liu2020simultaneously} &   \underline{86.8}     &   87.5    &      87.9    &   88.5    &   89.8    &   90.2    \\
HRNet    \cite{liu2020simultaneously} &   84.3     &   89.1    &      89.9    &   90.2    &   90.7    &   91.1    \\    
3DCNN \cite{luo2021intelligent}       &   83.5     &   86.2    &      89.5    &   90.3    &   92.4    &   93.2    \\
ViTPose                               &   \textbf{90.6}     &   \textbf{90.9}    &      \underline{91.4}    &   91.9    &   92.9    &   93.6    \\
T-ViTPose (\textbf{Ours}, $C$=~4)     &   N/A      &   89.3    &      89.9    &   \underline{92.4}    &   \underline{93.8}    &   \underline{94.5}    \\
T-ViTPose+MAE (\textbf{Ours}, $C$=~4) &   N/A      &   \underline{90.8}    &      \textbf{91.5}    &   \textbf{93.2}    &   \textbf{94.1}    &   \textbf{95.0}    \\
\Xhline{2\arrayrulewidth}
\end{tabularx}
}
\end{center}
\end{table}

Generally, pose estimation approaches perform significantly better given the temporal context \cite{li2022exploiting}. In our case, this is particularly seen on the carpet data where high levels of data ambiguity exist. As a result, we design our network to utilize \textit{temporal} crops throughout all stages of the network, thus improving performance and outperforming the 3DCNN solution \cite{luo2021intelligent} that uses temporal tiling and 3D convolutions only at later stages. Consequently, as shown in Table \ref{table:tb1}, our model outperforms other approaches, mostly on long time windows. For instance, T-ViTPose achieves third and fourth place when only using 4 and 8 frames on the Intelligent Carpet dataset, respectively. Similarly, in Table \ref{table:tb2}, On the SLP dataset, T-ViTPose achieves the second-best performance when using 4 frames but outperforms all benchmarks on longer time frames.

\begin{figure}
    \centering
    \includegraphics[width=1\columnwidth]{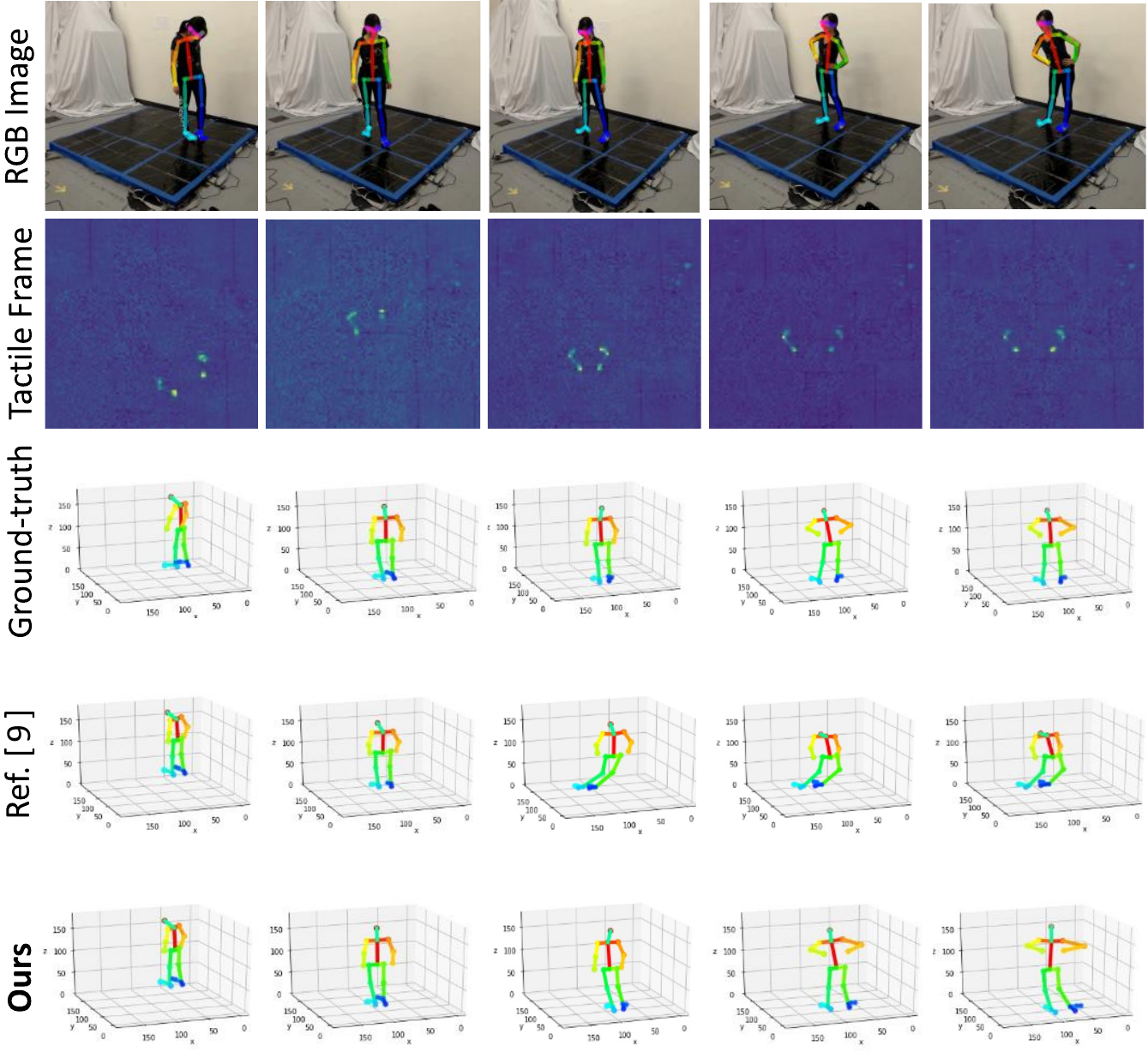}
    \caption{A comparison of our result and 3D-CNN \cite{luo2021intelligent}}
    \label{fig:fig2}
\end{figure}

Next, we conduct a parameter test to investigate the effect of temporal cropping on the performance of T-ViTPose in Table \ref{table:tb3}. We show that increasing the number of temporal patches consistently improves the performance. For instance, applying $4$ crops instead of $2$ reduces the error by $0.9cm$ and $3.9cm$ when $4$ and $32$ temporal frames are available. Furthermore, we show that using self-supervised pre-training can reduce the error by an average of $0.45cm$ across all conditions. Finally, we illustrate some examples of our method in Figure \ref{fig:fig2}, and compare our results against previous works, where we observe more accurate pose estimation by our model.

\begin{table}
\centering
\begin{center}
\caption{Effect of number of temporal crops and SSL pre-training on the performance of T-ViTPose.} 
\label{table:tb3}
\setlength
\tabcolsep{3pt}
\footnotesize
\centering
\resizebox{1\columnwidth}{!}{
\begin{tabularx}{\columnwidth}{c|c|YY}
\Xhline{2\arrayrulewidth}
\multirow{2}{*}{\textbf{SSL Pre-training}}   & \multirow{2}{*}{\textbf{Temporal Crops}}  & \multicolumn{2}{c}{\textbf{Frames}}        \\ 
\cline{3-4}
                &           &     4        &    32 \\ 
\Xhline{2\arrayrulewidth}
\cmark      &   1       &     29.8       & 23.4 \\ 
\cmark      &   2       &     29.3       & 20.4 \\ 
\cmark      &   4       &     28.4       & 16.5 \\ 
\Xhline{1\arrayrulewidth}
\xmark          &   1       &     30.2       & 23.1 \\ 
\xmark          &   2       &     30.1       & 21.4 \\ 
\xmark          &   4       &     28.9       & 16.9 \\ 
\Xhline{2\arrayrulewidth}
\end{tabularx}
}
\end{center}
\end{table}

\begin{table}[ht!]
\footnotesize
\begin{center}
\caption{Number of parameters, FLOPS, and inference time.} 
\label{table:tb4}
\setlength
\tabcolsep{3pt}
\centering
\resizebox{1\columnwidth}{!}{
\begin{tabularx}{\columnwidth}{c|YYc}
\Xhline{2\arrayrulewidth}
\textbf{Method} &  \textbf{FLOPS (G)} & \textbf{Parameters (M)} & \textbf{Inference Time (ms)} \\ 
\Xhline{2\arrayrulewidth}
ResNet18           &   0.72    & 15.52  & 79.23     \\
ResNet50           &   1.2     & 34.15  & 182.93    \\
ResNet101          &   1.88    & 53.14  & 357.11    \\
UNet               &   3.88    & 16.84  & 51.67     \\
HRNet              &   0.69    & 9.32   & 366.36    \\    
3DCNN              &   57.31   & 68.86  & 67.02     \\
ViTPose            &   3.52    & 93.31  & 383.53    \\
\textbf{Ours} T=1  &   3.57    & 93.3   & 410.45    \\
\textbf{Ours} T=2  &   6.73    & 94.51  & 413.00    \\
\textbf{Ours} T=4  &   13.07   & 99.88  & 401.24    \\
\Xhline{2\arrayrulewidth}
\end{tabularx}
}
\end{center}
\end{table}

Finally, Table \ref{table:tb4} provides the model parameters, inference time, and FLOPS for a batch of 20-frame input window. We show that our model has the same order of parameters as other top-performing approaches, is only 30 ms slower than HRNet or ViTPose, and uses 78\% less FLOPS than the previous state-of-the-art pressure-based pose estimation approach. Additionally, the training time of all models was 30 minutes per epoch with a batch size of 128, except 3DCNN, where training took 2 hours per epoch with a batch size of 64 due to the slow back-propagation caused by the large number of parameters from the tiling operation and the 3D convolution layers. These measurements were taken on an average of 1000 forward passes given a 96 $\times$ 96 array on an Nvidia 1080 Xp GPU.

\section{Conclusion}
\label{sec:conc}
In this paper, we presented a 3-stage solution for accurate 3D pose estimation from a temporal window of ambiguous pressure data. Specifically, we proposed a temporal variation of ViT by using $(2+1)D$ convolutions as the initial block, and pre-trained the network using the self-supervised masked auto-encoder strategy. After a few epochs of warm-up for modified modules, we trained our model using conventional 3D pose estimation objectives. We show that each element in our design significantly improves the prediction error over prior works in our experiments and set a new state of the art on two large-scale pressure mapping datasets.

\vspace{10pt}

\noindent \textbf{Acknowledgement.} This project was funded in part by Natural Sciences and Engineering Research Council of Canada.

\bibliographystyle{IEEEbib}
\bibliography{refs}

\end{document}